\documentclass[runningheads]{llncs}
\usepackage{graphicx}
\usepackage{silence}
\usepackage{tikz}
\usepackage{comment}
\usepackage{amsmath,amssymb} 
\usepackage{color}

\usepackage[accsupp]{axessibility}  

\usepackage[width=122mm,left=12mm,paperwidth=146mm,height=193mm,top=12mm,paperheight=217mm]{geometry}

\begin{document}
\pagestyle{headings}
\mainmatter
\def\ECCVSubNumber{100}  

\title{An Ensemble Approach for Multiple Emotion Descriptors Estimation Using Multi-task Learning} 

\titlerunning{An Ensemble Approach for Multiple Emotion ...} 
\authorrunning{Irfan Haider, Minh-Trieu Tran, ...} 
\author{Irfan Haider, Minh-Trieu Tran, Soo-Hyung Kim, Hyung-Jeong Yang, Guee-Sang Lee}
\institute{Chonnam National University}

\maketitle

\begin{abstract}

This paper illustrates our submission method to the fourth Affective Behavior Analysis in-the-Wild (ABAW) Competition. The method is used for the Multi-Task Learning Challenge. Instead of using only face information, we employ full information from a provided dataset containing face and the context around the face. We utilized the InceptionNet V3 model to extract deep features then we applied the attention mechanism to refine the features. After that, we put those features into the transformer block and multi-layer perceptron networks to get the final multiple kinds of emotion. Our model predicts arousal and valence, classifies the emotional expression and estimates the action units simultaneously. The proposed system achieves the performance of 0.917 on the MTL Challenge validation dataset.

\end{abstract}

\section{Introduction}

Emotion recognition plays an important role in many researches because of its multiple application  in different fields such as autonomous driving, or medicine. Emotion Recognition is also significant in the human effective behaviour research and it has been studied many times with the development of big data and deep learning technologies. Emotion can be detected by several ways such as images, videos, speech and by using bio- signals. Human facial expressions are not enough to recognise the emotion by images or videos, scene context also effective to rather than only using the facial features\cite{fabian2016emotionet}. Meanwhile, people recognize human emotions not only from facial features but also from their surroundings, such as movement, communication with others, and location\cite{barrett2011context}\cite{aminoff2013role}.

Traditional recognition studies only focused on facial features and ignore the other information i.e., contextual information that is very effective in recognising the human emotion accurately. Previously, due to a lack of vast amounts of data collected in actual scenarios, research focused on controlled scenarios\cite{kollias2022abaw}.
However, digital networks and social media have previously become broadly used and a huge amount of data has become available.

Human emotions have recently been the focus of psychology. The most common type of emotion representation is deterministic, which includes the seven basic categories of anger, disgust, fear, happiness, sadness, surprise, and neutrality\cite{ekman2003darwin}. 

Emotion recognition research is booming in computer vision. Continuous research in this field and development in deep learning it is gaining more and more attention. The proof is available as large number of datasets Aff-Wild\cite{zafeiriou2017aff}\cite{kollias2017recognition}, s-Aff-Wild2\cite{kollias2021affect}\cite{kollias2020deep}\cite{kollias2020va}\cite{kollias2019expression}\cite{kollias2019deep}\cite{kollias2018photorealistic}, Aff-Wild2\cite{kollias2018aff}.

Some problems related to emotion recognition are discussed in 3rd Affective Behavior Analysis in-the wild (ABAW) Competition is held in conjunction with IEEE International Conference on Computer Vision and
Pattern Recognition (CVPR), 2022. 3rd ABAW includes
Valence-Arousal (VA) Estimation Challenge, Expression
(Expr) Classification Challenge, Action Unit (AU) Detection
Challenge, Multi-Task-Learning (MTL) Challenge. Many teams participate every time in ABAW\cite{kollias2022abaw4} challenge to improve the performance of human emotion models in the real world.

\section{Related Work}

For many years, researchers have researched about human emotion recognition utilizing various forms of human emotion, such as fundamental facial expressions, action units, and valence arousal. Mostly researchers focused on the facial features to recognize the human emotion\cite{fabian2016emotionet}\cite{li2018occlusion}. Some methods are based on the facial action system\cite{ekman1978facial}\cite{eleftheriadis2014discriminative}, which uses a set of local facial postures to encrypt facial expression. The majority of studies have depended on analyzing human faces, which restricts their capability to utilise semantic features for emotion recognition in the wild\cite{kollias2021distribution}. 

Many researchers have researched about how to improve recognising the emotion and to improve the performance, they used different inputs. Some used only images\cite{deng2020multitask}\cite{youoku2020multi}\cite{youoku2021multi} and some used audio data\cite{deng2021iterative}\cite{kuhnke2020two} with images. Different researches used different approaches\cite{hasani2017facial} for classifying and recognising the human emotions. In emotion recognition different approaches has been used for classifying the emotions. Deep Convolutional network(CNN) is used in different fields of research\cite{krizhevsky2012imagenet}\cite{simonyan2014very}. Especially CNN have big contribution in the computer vision field. After having good performance\cite{zhou2016learning} in computer vision, CNN is investigated to classifying the facial emotions\cite{singh2016track}.

In our study we used CNN due to its better performance and we used Inception-V3\cite{zou2014chronological} model that is a deep CNN architecture. This model is proposed by  Szegedy et al. in the Large-Scale ImageNet Visual Identification Challenge 2014. It's goal was to improve computational performance and reduce the parameters in applications. Inception-V3 model is faster than VGG. Basically it is the modification of AlexNet. Last year top 1st ranked author of ABAW  challenge also used the same model in their work and achieved good results\cite{deng2022multiple}.

\begin{figure*}
\centering
\includegraphics[height=17cm]{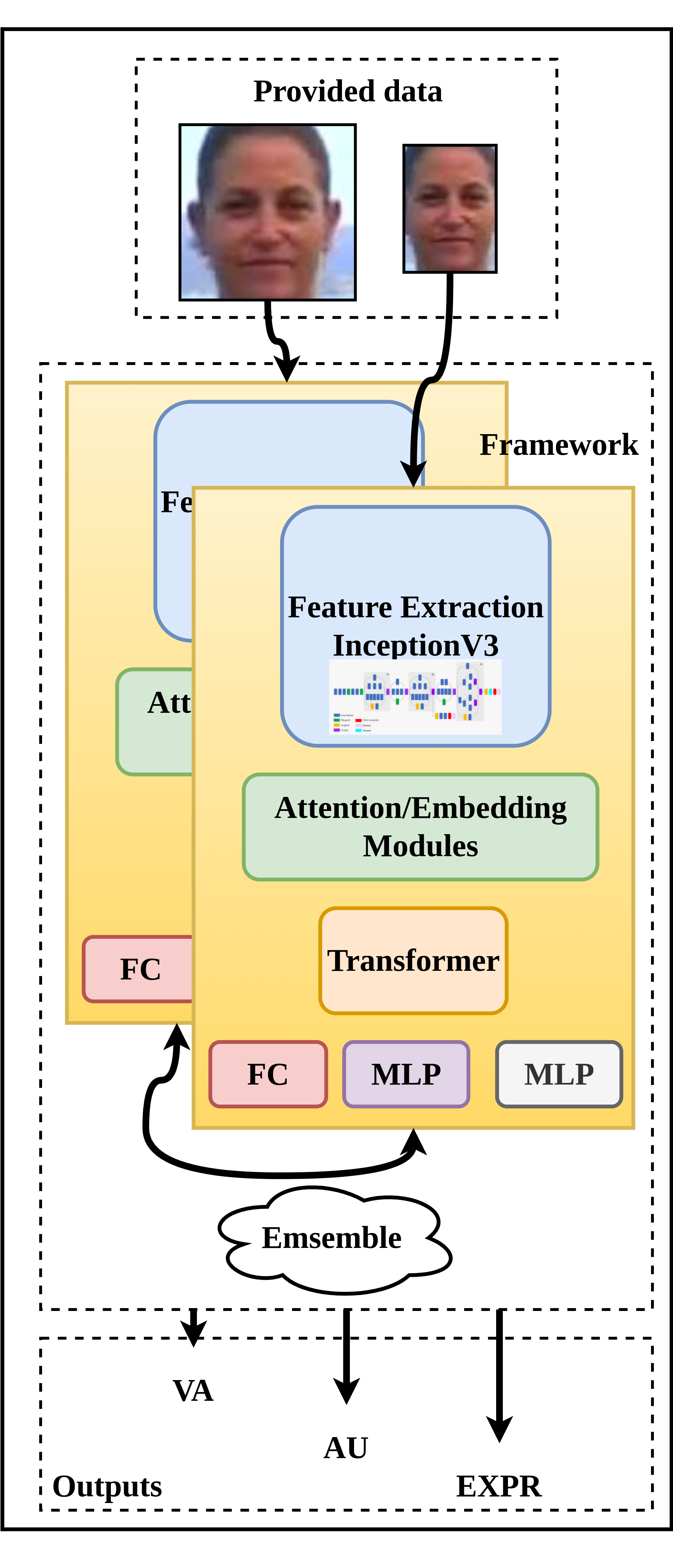}
        \caption{Illustration of our proposed architecture.}
        \label{fig1}
\end{figure*}

\section{Proposed Method}      
Our proposed architecture has been shown in figure\ref{fig1}. We employ two kinds of information from the provided dataset: cropped and cropped aligned images. Cropped images include the facial, while the cropped aligned images contain the facial and also the context around. Based on the results of the first ranking last year\cite{deng2022multiple}, we utilize the model and add some modifications. Instead of using only the cropped aligned images, we explored full information from the provided dataset. Basically, our approach is based on three main parts: feature extraction, attention mechanism, and classifiers.

For the first part, we used Inception Net V3 to extract the features from the input image. 
Inception v3 is a deep network for the image recognition field. The model itself contains symmetric and asymmetric building blocks with integrated convolutions and pooling layers inside. This model shows a superb result with higher than 78 \% accuracy on the ImageNet dataset. The feature extractor gives the feature output with the sizes of 768 $\times$ 17 $\times$ 17.

In the second part, we will mention the mechanism modules that help the model to focus better on meaningful regions. After having the feature maps, we  put them through the attention module to learn the spatial attention for each region. Based on the combination of attention maps and features, we have the embedding feature vector of each region. 

The final part included transformer block and multi-layer perception networks. We apply the transformer to learn the correlation between different regions related to different action units. After having the output features from the transformer block, we put them into action unit classifiers. Note that each action unit has a separate classifier. Besides the action unit classifiers, we proposed the valence/arousal predictors and expression classifier. For the valence/arousal predictor, we use two fully connected layers. We employ a multi-layer perception with three fully connected layers for the expression classifier.

During training, we try multiple types of loss functions. Particularly, in the action unit prediction and emotional expression classification, we use the cross-entropy loss function. With the arousal/valence prediction task, a negative Concordance Correlation Coefficient (CCC) is employed as a loss function. Finally, the total loss is the sum of three elemental losses.

We use several kinds of evaluation metrics in this paper. Firstly, the F1 score metric is used in the evaluation step to compute the performance of action unit prediction and emotional expression classification tasks. Secondly, the performance of valence/arousal prediction is calculated by CCC metric. The total metric for multi-task prediction is based on the combination of the F1 score and CCC metrics.
\section{Experimental Results}  
The proposed method is trained on the system with only one Nvidia GTX 3090 GPU card, 24GB memory, and with SGD optimizer. We trained the model for 50 epochs, batch size 24, and the learning rate of 0.001. Note that we train in the same number of epochs for a fair comparison between the proposed method and last year's first rank method. 

Table \ref{table1} shows comparison results between our proposed method with the top-ranking ABAW3 method in the EXPR classification task. The results prove two things. 

First is the meaning of context in emotion recognition which contains faces can bring more useful information for the EXPR classification task than only the face information. It is similar to AU classification and Valence prediction tasks, shown in Tables \ref{table2}, \ref{table3}, respectively. However, for the Arousal prediction task results, shown in Table \ref{table4}, information that comes from the face only seems more meaningful than the combination of face and context around.

The second thing is that the results show the ensemble approach brings better results than using only a separate model. 

\begin{figure*}
\centering
\includegraphics[height=5cm]{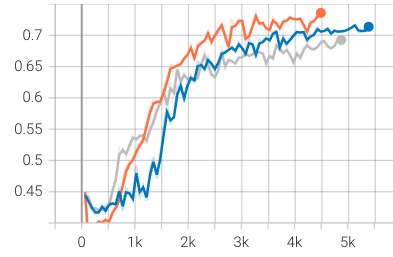}
        \caption{Illustration of accuracy in action unit prediction during training steps. The orange line indicates the performance of training from facial (cropped images). The blue line indicates the performance of training from facial and context around (cropped aligned images). The grey line indicates the performance of training from last year's first ranking provided source code from facial and context around.}
        \label{fig2}
\end{figure*}

\begin{figure*}
\centering
\includegraphics[height=5cm]{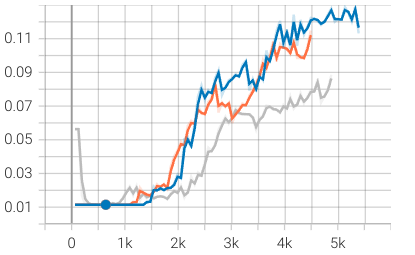}
        \caption{Illustration of F1 score in expression classification during training steps. The orange line indicates the performance of training from facial (cropped images). The blue line indicates the performance of training from facial and context around (cropped aligned images). The grey line indicates the performance of training from last year's first ranking provided source code from facial and context around.}
        \label{fig3}
\end{figure*}

\begin{figure*}
\centering
\includegraphics[height=5cm]{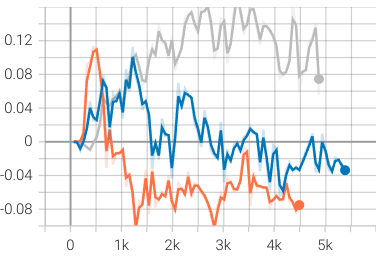}
        \caption{Illustration of Concordance Correlation Coefficient in arousal prediction during training steps. The orange line indicates the performance of training from facial (cropped images). The blue line indicates the performance of training from facial and context around (cropped aligned images). The grey line indicates the performance of training from last year's first ranking provided source code from facial and context around.}
        \label{fig4}
\end{figure*}

\begin{figure*}
\centering
\includegraphics[height=5cm]{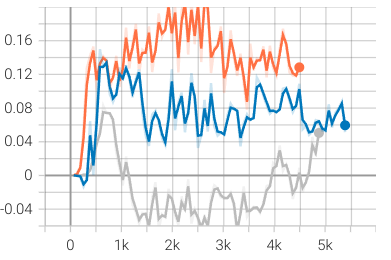}
        \caption{Illustration of Concordance Correlation Coefficient in valence prediction during training steps. The orange line indicates the performance of training from facial (cropped images). The blue line indicates the performance of training from facial and context around (cropped aligned images). The grey line indicates the performance of training from last year's first ranking provided source code from facial and context around.}
        \label{fig5}
\end{figure*}

\begin{figure*}
\centering
\includegraphics[height=6cm]{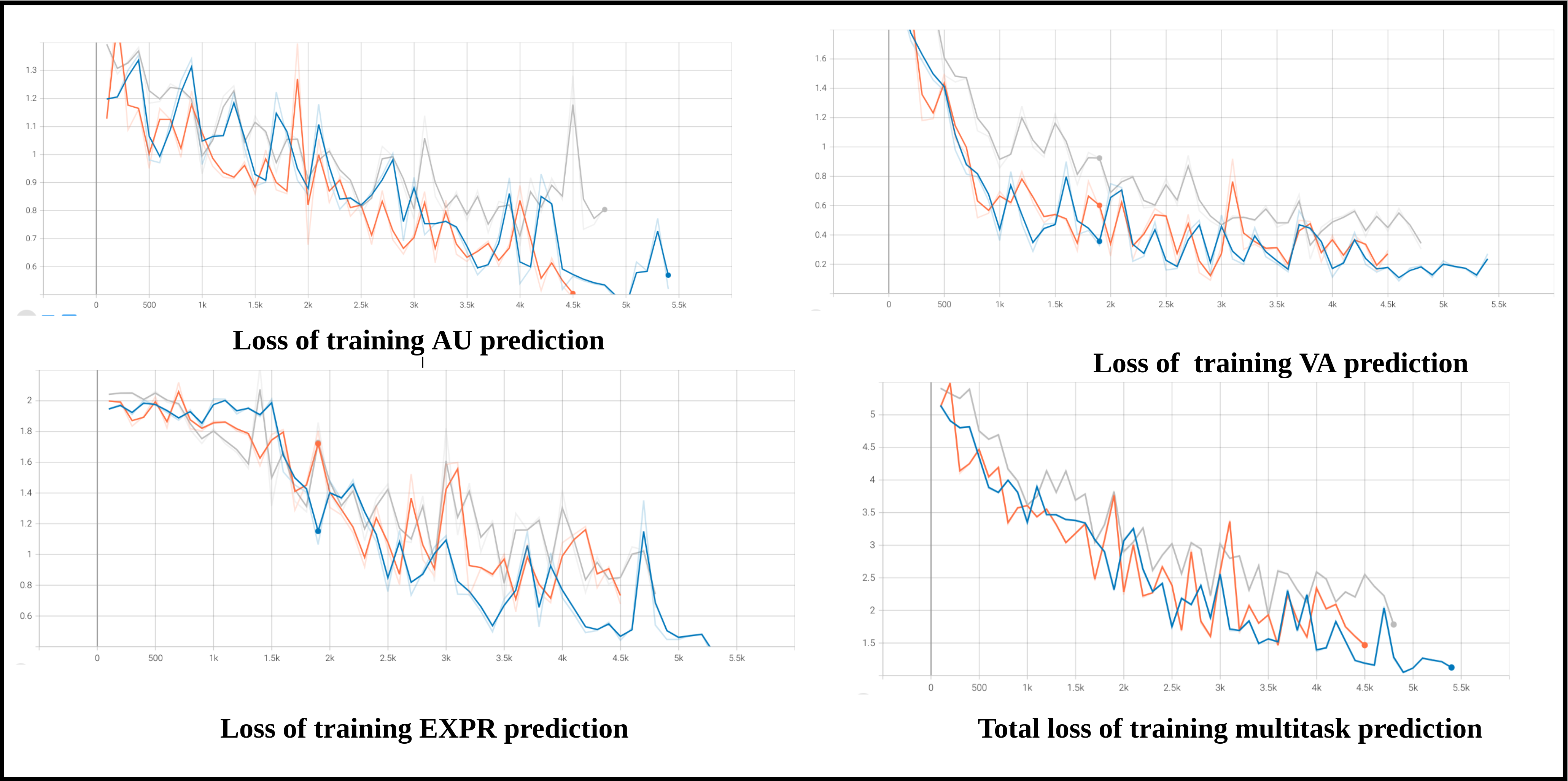}
        \caption{Illustration loss reduction during training steps. The orange line indicates the performance of training from facial (cropped images). The blue line indicates the performance of training from facial and context around (cropped aligned images). The grey line indicates the performance of training from last year's first ranking provided source code from facial and context around.}
        \label{fig6}
\end{figure*}

The information of the training procedure are presented in Figures \ref{fig2},\ref{fig3},\ref{fig4},\ref{fig5},\ref{fig6}. Figure \ref{fig2} is the illustration of accuracy in action unit prediction during training steps. Figure \ref{fig3} is the illustration of the F1 score in expression classification during training steps. Figures \ref{fig4}, \ref{fig5}  is the illustration of the Concordance Correlation Coefficient in Arousal and Valence predictions during training steps. Fig \ref{fig6} presents loss reduction during training steps. From those images, the orange line indicates the performance of training from facial (cropped images). The blue line indicates the performance of training from facial and context around (cropped aligned images). The grey line indicates the performance of training from last year’s first ranking provided source code from facial and context around.

\begin{table}[ht!]
\centering
    \caption{The comparison results between our proposed method with the top-ranking ABAW3 method in the EXPR classification task.}
    \label{table1}
\begin{tabular}{l|c}
\hline
Method                 & F1 Score \\ \hline
Last year 1st rank     & 0.1023        \\
Ours (Cropped)         & 0.1692        \\
Ours (Cropped Aligned) & 0.2194        \\
Ours (Ensemble)        & 0.3055        \\ \hline
\end{tabular}
\end{table}

\begin{table}[ht!]
\centering
    \caption{The comparison results between our proposed method with the top-ranking ABAW3 method in AU classification task.}
    \label{table2}
\begin{tabular}{l|c}
\hline
Method                 & F1 Score \\ \hline
Last year 1st rank     & 0.4714        \\
Ours (Cropped)         & 0.4856        \\
Ours (Cropped Aligned) & 0.4860        \\
Ours (Ensemble)        & 0.4939        \\ \hline
\end{tabular}
\end{table}

\begin{table}[ht!]
\centering
    \caption{The comparison results between our proposed method with the top-ranking ABAW3 method in the Valence Prediction task.}
    \label{table3}
\begin{tabular}{l|c}
\hline
Method                 & CCC \\ \hline
Last year 1st rank     & 0.0534        \\
Ours (Cropped)         & 0.0520        \\
Ours (Cropped Aligned) & 0.1091        \\
Ours (Ensemble)        & 0.1597        \\ \hline
\end{tabular}
\end{table}

\begin{table}[ht!]
\centering
    \caption{The comparison results between our proposed method with the top-ranking ABAW3 method in the Arousal Prediction task.}
    \label{table4}
\begin{tabular}{l|c}
\hline
Method                 & CCC \\ \hline
Last year 1st rank     & 0.0504        \\
Ours (Cropped)         & 0.0735        \\
Ours (Cropped Aligned) & 0.0446        \\
Ours (Ensemble)        & 0.0754        \\ \hline
\end{tabular}
\end{table}

\begin{table}[ht!]
\centering
    \caption{The comparison results between our proposed method with baseline and the top-ranking ABAW3 method in the Multi-Task Prediction.}
    \label{table5}
\begin{tabular}{l|c}
\hline
Method                 & Total Metric \\ \hline
Baseline               & 0.3000       \\
Last year 1st rank     & 0.6256        \\
Ours                   & 0.9170        \\ \hline
\end{tabular}
\end{table}

\section{Conclusion}      
This paper presents an ensemble approach for multiple emotion descriptors estimation using multi-task learning. The context is one of the important factors in improving the performance of emotion recognition tasks. Additionally, Instead of using a separate framework, the ensemble approach still can bring more benefit to recognition issues. Our code implementation is available at https://github.com/tmtvaa/abaw4.

\section{Acknowledgement}     
This research was supported by the Bio \& Medical Technology Development Program of the National Research Foundation (NRF) \& funded by the Korean government (MSIT) (NRF-2019M3E5D1A02067961) and by Basic Science Research Program through the National Research Foundation of Korea(NRF) funded by the Ministry of Education(NRF-2018R1D1A3B05049058 \& NRF-2020R1A4A1019191).

\bibliographystyle{splncs04}
\bibliography{egbib}
\end{document}